\documentclass[11pt,a4paper,dvipsnames]{article}
\usepackage{times,latexsym}
\usepackage{url}
\usepackage[T1]{fontenc}
\usepackage{todonotes}
\usepackage{arydshln}

\usepackage{hyperref}
\usepackage{tabularx}
\usepackage{url}
\usepackage{booktabs}
\usepackage{multicol, multirow}
\usepackage{wrapfig}
\usepackage{subfigure}
\usepackage{enumitem}
\usepackage{color,soul}
\usepackage{longtable}
\usepackage{multirow} 
\usepackage{tabularx}
\usepackage{ragged2e}  
\newcolumntype{Y}{>{\RaggedRight\arraybackslash}X} 
\usepackage{xcolor}
\usepackage{color, colortbl,soul}
\definecolor{LGray}{gray}{0.9}
\definecolor{Gray}{gray}{0.8}
\definecolor{DGray}{gray}{0.7}
\sethlcolor{Gray}
\setuldepth{Berlin}

\usepackage[acceptedWithA]{tacl2021v1}

\usepackage{xspace,mfirstuc,tabulary}

\newif\iftaclinstructions
\taclinstructionsfalse 
\iftaclinstructions
\renewcommand{\confidential}{}https://www.overleaf.com/project/60f592b65d2e4f3f6b6b568e
\renewcommand{\anonsubtext}{(No author info supplied here, for consistency with
TACL-submission anonymization requirements)}
\newcommand{\instr}
\fi

\iftaclpubformat 

\else

\fi

\title{Template-based Abstractive Microblog Opinion Summarisation}
\author{
 Iman Munire Bilal$^{1,4}$, Bo Wang$^{2,4}$, Adam Tsakalidis$^{3,4}$, Dong Nguyen$^{5}$, Rob Procter$^{1,4}$, Maria Liakata$^{1,3,4}$ \\
$^1$ Department of Computer Science, University of Warwick \\
$^2$ Center for Precision Psychiatry, Massachusetts General Hospital \\
$^3$ School of Electronic Engineering and Computer Science, Queen Mary University of London\\
$^4$ The Alan Turing Institute, London, UK\\
$^5$ Department of Information and Computing Sciences, Utrecht University\\
   \texttt{$\{$iman.bilal|rob.procter$\}$@warwick.ac.uk bwang29@mgh.harvard.edu}\\

  \texttt{
  $\{$atsakalidis|mliakata$\}$@qmul.ac.uk d.p.nguyen@uu.nl}\\}

\date{}

\begin{document}
\maketitle
\begin{abstract}
We introduce the task of microblog opinion summarisation (MOS) and share a dataset of 3100 gold-standard opinion summaries to facilitate research in this domain. The dataset contains summaries of tweets spanning a 2-year period and covers more topics than any other public Twitter summarisation dataset. Summaries are abstractive in nature and have been created by journalists skilled in summarising news articles following a template separating factual information (main story) from author opinions. Our method differs from previous work on generating gold-standard summaries from social media, which usually involves selecting representative posts and thus favours extractive summarisation models. To showcase the dataset's utility and challenges, we benchmark a range of abstractive and extractive state-of-the-art summarisation models and achieve good performance, with the former outperforming the latter. We also show that fine-tuning is necessary to improve performance and investigate the benefits of using different sample sizes.
\end{abstract}

\section{Introduction}
Social media has gained prominence as a means for the public to exchange opinions on a broad range of topics. Furthermore, its social and temporal properties make it a rich resource for policy makers and organisations to track public opinion on a diverse range of issues \citep{procter2013, chou2018addressing, kalimeri2019human}. 
However, understanding opinions about different issues and entities discussed in large volumes of posts in platforms such as Twitter is a difficult task. Existing work on Twitter employs extractive summarisation \citep{inouye, zubiaga2012realtime, wang2017totemss, stance_summ} to filter through information by ranking and selecting tweets according to various criteria. However, this approach unavoidably ends up including incomplete or redundant information \citep{rotten_tomatoes}.
\begin{table}[!h]
    \scriptsize
    \begin{tabular}{|p{7.5cm}|}
        \hline
        \textbf{Human Summary}\\
        \hline
        \multirow{4}{7.5cm}{\textcolor{blue}{\textbf{Main Story:} The UK government faces intense backlash after its decision to fund the war in Syria.} \textcolor{red}{\textbf{Majority Opinion:} The majority of users criticise UK politicians for not directing their efforts to more important domestic issues like the NHS, education and homelessness instead of the war in Syria.} \textcolor{teal}{\textbf{Minority Opinion:} Some users accuse the government of its intention to kill innocents by funding the war.}}\\
        \\ \\ \\ \\ \\ \\
        \hline
        \textbf{Tweet Cluster}\\
        \hline
        \multirow{3}{7.5cm}{\textcolor{red}{It is shocking to me how the NHS is on its knees and the amount of homeless people that need help in this country}...\textcolor{blue}{but we have funds for war!..SAD}}
        \\ \\ \\
        \multirow{3}{7.5cm}{\textcolor{red}{The government cannot even afford to help the homeless people of Britain} yet \textcolor{blue}{they can afford to fund a war?} It makes no proper sense at all}
        \\ \\ \\
        \multirow{3}{7.5cm}{\textcolor{teal}{They spend so much on sending missiles to murder innocent people} and they complain daily about \textcolor{red}{homeless on the streets? Messed up.}}
        \\ \\ \\
        \multirow{2}{7.5cm}{Also, \textcolor{red}{no money to resolve the issues of the homeless or education or the NHS.} \textcolor{blue}{Yet loads of money to drop bombs? \#SyriaVote}}
        \\ \\ 
        \hline
    \end{tabular}
    \caption{Abridged cluster of tweets and its corresponding summary. Cluster content is color-coded to represent information overlap with each summary component: \textcolor{blue}{blue} for Main Story,  \textcolor{red}{red} for Majority Opinion and  \textcolor{teal}{green} for Minority Opinion}
    \label{TOS_example}
\end{table}
\normalsize

To tackle this challenge we introduce \textit{Microblog opinion summarisation} (MOS), which we define as a multi-document summarisation task aimed at capturing diverse reactions and stances (opinions) of social media users on a topic. While here we apply our methods to Twitter data readily available to us, we note that this summarisation strategy is also useful for other microblogging platforms. An example of a tweet cluster and its opinion summary is shown in \textit{Table} \ref{TOS_example}. As shown, our proposed summary structure for MOS separates the factual information (story) from reactions to the story (opinions); the latter is further divided according to the prevalence of different opinions. We believe that making combined use of stance identification, sentiment analysis and abstractive summarisation is a challenging but valuable direction in aggregating opinions expressed in microblogs. 

The availability of high quality news article datasets has meant that recent advances in text summarisation have focused mostly on this type of data \citep{nallapati2016abstractive,grusky-etal-2018-newsroom,fabbri-etal-2019-multi,gholipour-ghalandari-etal-2020-large}. Contrary to news article summarisation, our task focuses on summarising an event as well as ensuing public opinions on social media. 
Review opinion summarisation \citep{ganesan-etal-2010-opinosis,angelidis-lapata-2018-summarizing} is related to MOS and faces the same challenge of filtering through large volumes of user-generated content. While recent work \citep{Chu2019MeanSumAN, brazinskas-etal-2020-unsupervised} aims to produce review-like summaries that capture the consensus, MOS summaries 
inevitably include a spectrum of stances and reactions. In this paper we make the following contributions:
\begin{enumerate}[itemsep=1.5pt,parsep=1.5pt]
    \item We introduce the task of microblog opinion summarisation (MOS) and provide detailed guidelines. 
    \item We construct a corpus\footnote{This is available at \url{https://doi.org/10.6084/m9.figshare.20391144}} of tweet clusters and corresponding multi-document summaries produced by expert summarisers following our detailed guidelines. 
    \item We evaluate the performance of existing state-of-the-art models and baselines from three summarisation domains (news articles, Twitter posts, product reviews) and four model types (abstractive vs. extractive, single document vs. multiple documents) on our corpus, showing the superiority of neural abstractive models. We also investigate the benefits of fine-tuning with various sample sizes.
\end{enumerate}

\normalsize
\section{Related Work}
\label{Sec_2}
\textbf{Opinion Summarisation} has focused predominantly on customer reviews with datasets spanning reviews on Tripadvisor \citep{ganesan-etal-2010-opinosis}, Rotten Tomatoes \citep{rotten_tomatoes}, Amazon \citep{amazon_reviews, angelidis-lapata-2018-summarizing} and Yelp (Yelp Dataset Challenge)\citep{yelpchallenge}. 

Early work~\citet{ganesan-etal-2010-opinosis} prioritised redundancy control and concise summaries. More recent approaches \citep{angelidis-lapata-2018-summarizing, amplayo-lapata-2020-unsupervised, angelidis-etal-2021-extractive, isonuma2021unsupervised} employ aspect driven models to create relevant topical summaries. While product reviews have a relatively fixed structure, MOS operates on microblog clusters where posts are more loosely related, which poses an additional challenge. Moreover, while the former generally only encodes the consensus opinion \citep{brazinskas-etal-2020-unsupervised, Chu2019MeanSumAN}, our approach includes both majority and minority opinions. 

\noindent \textbf{Multi-document summarisation} has gained traction in non-opinion settings and for news events in particular. DUC \citep{DUC} and TAC conferences pioneered this task by introducing datasets of 139 clusters of articles paired with multiple human-authored summaries. Recent work has seen the emergence of larger scale datasets such as WikiSum \citep{Liu2018GeneratingWB}, Multi-News \citep{fabbri-etal-2019-multi} and WCEP \citep{gholipour-ghalandari-etal-2020-large} to combat data sparsity. Extractive \citep{wang-etal-2020-heterogeneous,wang-etal-2020-spectral,liang-etal-2021-improving} and abstractive \citep{jin-etal-2020-multi} methods have followed from these multi-document news datasets.

\noindent \textbf{Twitter Summarisation} is recognised by \citet{cao2016tgsum} to be a promising direction for tracking reaction to major events. As tweets are inherently succinct and often opinionated \citep{mohammad-etal-2016-semeval}, this task is at the intersection of multi-document and opinion summarisation. The construction of datasets \citep{nguyen-etal-2018-tsix, ijcai2017-0581} usually requires a clustering step to group tweets together under specific temporal and topical constraints, which we include within our own pipeline. Work by \citet{stance_summ} and \citet{football} makes use of the subjective nature of tweets by identifying two stances for each topic to be summarised; we generalise this idea and do not impose a restriction on the number of possible opinions on a topic.
The lack of an abstractive gold standard means that the majority of existing Twitter models are extractive \citep{Alsaedi_Burnap_Rana_2021, inouye, stance_summ, football}. Here we provide such an abstractive gold standard and show the potential of neural abstractive models for microblog opinion summarisation. 

\section{Creating the MOS Dataset}
\subsection{Data Sources}
Our MOS corpus consists of summaries of microblog posts originating from two data sources, both involving topics that have generated strong public opinion: \textbf{COVID-19} \citep{coviddata:1} and \textbf{UK Elections} \citep{our_paper}. 
\begin{itemize}
    \item \textbf{COVID-19}: \citet{coviddata:1} collected tweets by tracking COVID-19 related keywords (e.g., \textit{coronavirus}, \textit{pandemic}, \textit{stayathome}) and accounts (e.g., \textit{@CDCemergency}, \textit{@HHSGov}, \textit{@DrTedros}). We use data collected between January 2020 and January 2021 which at the time was the most complete version of this dataset.
    \vspace{-0.2cm}
    \item  \textbf{UK Elections}: The \textbf{Election} dataset consists of all geo-located UK tweets posted between May 2014 and May 2016. The tweets were filtered using a list of 438 election-related keywords and 71 political party aliases curated by a team of journalists.
\end{itemize}

\noindent We follow the methodology in \newcite{our_paper} to obtain opinionated, coherent clusters of between 20--50 tweets: the clustering step employs the GSDMM-LDA algorithm \citep{gsdmmlda:1}, followed by thematic coherence evaluation \cite{our_paper}. The latter is done by aggregating exhaustive metrics BLEURT \citep{sellam-etal-2020-bleurt}, BERTScore \citep{bert-score} and TF-IDF to construct a random forest classifier to identify coherent clusters. Our final corpus is created by randomly sampling 3100 clusters\footnote{Limited resources available for annotation determined the size of the MOS corpus.}, 1550 each from the COVID-19 and Election datasets.

\subsection{Summary Creation}
The summary creation process was carried out in 3 stages on the Figure Eight platform by 3 journalists experienced in sub-editing. Following \citet{iskender-etal-2021-reliability}, a short pilot study was followed by a meeting with the summarisers to ensure the task and guidelines were well understood. Prior to this, the design of the summarisation interface was iterated to ensure functionality and usability (See \textit{Appendix A} for interface snapshots). 

In the first stage, the summarisers were asked to read a cluster of tweets and state whether the opinions within it could be easily summarised by assigning one of three cluster types:
\begin{enumerate}[itemsep=1.5pt,parsep=1.5pt]
\item \textbf{Coherent Opinionated}: there are clear opinions about a common main story expressed in the cluster that can be easily summarised.
\item \textbf{Coherent Non-opinionated}: there are very few or no clear opinions in the cluster, but a main story is clearly evident and can be summarised.
\item \textbf{Incoherent}: no main story can be detected. This happens when the cluster contains diverse stories to which no majority of tweets refers, hence it cannot be summarised.
\end{enumerate}

Following \newcite{our_paper} on thematic coherence, we assume a cluster is coherent, if and only if, its contents can be summarised. Thus, both Coherent Opinionated and Coherent Non-opinionated can be summarised, but are distinct with respect to the level of subjectivity in the tweets, while Incoherent clusters cannot be summarised.

In the second stage, \emph{information nuggets} are defined in a cluster as important pieces of information to aid in its summarisation. The summarisers were asked to highlight information nuggets when available and categorise their aspect in terms of: WHAT, WHO, WHERE, REACTION and OTHER. Thus, each information nugget is a pair consisting of the text and its aspect category (see \textit{Appendix A} for an example). Inspired by the pyramid evaluation framework \citep{nenkova2004evaluating} and extractive-abstractive two-stage models in the summarisation literature \citep{lebanoff-etal-2018-adapting, sum_situational_tweets, Liu2018GeneratingWB}, information nuggets have a dual purpose: (1) helping summarisers create the final summary and (2) constituting an extractive reference for summary informativeness evaluation (See \ref{non_FT_guidelines}). 

In the third and final stage of the process, the summarisers were asked to write a short template-based summary for coherent clusters. Our chosen summary structure diverges from current summarisation approaches that reconstruct the “most popular opinion” \citep{brazinskas-etal-2020-unsupervised,angelidis-etal-2021-extractive}. Instead, we aim to showcase a spectrum of diverse opinions regarding the same event. Thus, the summary template comprises three components: \textit{Main Story}, \textit{Majority Opinion}, \textit{Minority Opinion(s)}. The component \textit{Main Story} serves to succinctly present the focus of the cluster (often an event), while the other components describe opinions about the main story. Here, we seek to distinguish the most popular opinion (\textit{Majority opinion}) from ones expressed by a minority (\textit{Minority opinions}). This structure is consistent with the work of \citet{gerani-etal-2014-abstractive} in template-based summarisation for product reviews, who quantify the popularity of user opinions in the final summary.

For ``Coherent Opinionated clusters'', summarisers were asked to identify the majority opinion within the cluster and if it exists, to summarise it, along with any minority opinions. If a majority opinion could not be detected, then the minority opinions were summarised. The final summary of ``Coherent Opinionated clusters'' is the concatenation of the three components: \textit{Main story} + \textit{Majority Opinion} (if any) + \textit{Minority Opinion(s)} (if any). In 43$\%$ of opinionated clusters in our MOS corpus a majority opinion and at least one minority opinion were identified. Additionally, in 12$\%$ of opinionated clusters, 2 or more main opinions were identified (See Appendix \ref{Appen_C}, Table \ref{tab:appen_example2}), but without a majority opinion as there is a clear divide between user reactions. For clusters with few or no clear opinions (Coherent Non-opinionated), the final summary is represented by the \textit{Main Story} component. Statistics regarding the annotation results are shown in \textit{Table} \ref{corpus_statistics}.
\begin{table}[!h]
    \centering
    \resizebox{.9\linewidth}{!}{%
    \begin{tabular}{cccc}
         \toprule
         & \textbf{Total} & \textbf{COVID-19} & \textbf{Election} \\
         \midrule
         \textbf{Size (\#clusters)} & 3100 & 1550 & 1550 \\
         \textbf{Coherent Opinionated} & 42\% & 
         41\% & 43\% \\
      
        \textbf{Coherent Non-opinionated} & 30\% & 24\% & 37\% \\
     
         \textbf{Incoherent} & 28\% & 35\% & 20\% \\
   
         \bottomrule
    \end{tabular}}
    \caption{Annotation statistics of our MOS corpus}
    \label{corpus_statistics}
\end{table}
\normalsize
\vspace{-0.4cm}
\subsubsection*{Agreement Analysis}
Our tweet summarisation corpus consists of 3100 clusters. Of these, a random sample of 100 clusters was shared among all three summarisers to compute agreement scores. Each then worked on 1000 clusters.

We obtain a Cohen's Kappa score of $\kappa = 0.46$ for the first stage of the summary creation process, which involves categorising clusters as either Coherent Opinionated, Coherent Non-opinionated or Incoherent. Previous work \citep{Feinstein1990HighAB} highlights a paradox regarding Cohen's kappa 
in that high levels of agreement do not translate to high kappa scores in cases of highly imbalanced datasets. In our data, at least 2 of the 3 summarisers agreed on the type of cluster in 97$\%$ of instances. 

In addition, we evaluate whether the concept of `coherence/summarisability' is uniformly assessed, i.e., we check whether summarisers agree on what clusters can be summarised (Coherent clusters) and which clusters are too incoherent. We find that 83 out of 100 clusters were evaluated as coherent by the majority, of which 65 were evaluated as uniformly coherent by all. 

ROUGE-1,2,L and BLEURT \citep{sellam-etal-2020-bleurt} are used as proxy metrics to check the agreement in terms of summary similarity produced between the summarisers. We compare the consensus between the complete summaries as well as individual components such as the main story of the cluster, its majority opinion and any minority opinions in Table \ref{tab:agr_sum}. The highest agreement is achieved for the Main Story, followed by Majority Opinion and Minority Opinions. These scores can be interpreted as upper thresholds for the lexical and semantic overlap later in Section \ref{Sec_6}.

\begin{table}[h]
    \centering
    \resizebox{.9\linewidth}{!}{%
    \begin{tabular}{lcccc}
         & $\textrm{R-1}_{f_1}$ & $\textrm{R-2}_{f_1}$ & $\textrm{R-L}_{f_1}$ & BLEURT \\
         \midrule
         \textbf{Summary} & 37.46 & 17.91& 30.16 & -.215\\
         \textbf{Main Story} & 35.15 & 12.98 & 34.59& -.324 \\
         \textbf{Majority Opinion} & 27.53 & 6.15 & 25.95  & -.497\\
         \textbf{Minority Opinion(s)} &22.90& 5.10 & 24.39 & -.703\\
         
    \end{tabular}}
    \caption{Agreements between summarisers wrt to final summary, main story, majority opinion and minority opinions using ROUGE-1,2,L and BLEURT.}
    \label{tab:agr_sum}
\end{table}

\subsection{Comparison with other Twitter datasets}

\begin{table*}[!h]
    \centering
    \resizebox{.9\textwidth}{!}{%
    \begin{tabular}{lcccccc}
         \textbf{Dataset} & \textbf{Time span} & \textbf{\#keywords} &\textbf{\#clusters} & \textbf{Avg. Cluster Size} & \textbf{Summary} & \textbf{Avg. Summary Length} \\
         &&&&(\#posts) && (\#tokens) \\
         \midrule
         COVID-19 & 1 year &41 &1003 & 31 &Abstractive & 42\\
         Election & 2 years &112 &1236 & 30 &Abstractive & 36 \\
         
         \citet{inouye} & 5 days & 50 &200&25&Extractive&17 \\
         SMERP \citep{smerp-ecir-2017} & 3 days&N/A & 8 &359 &Extractive &303\\
         TSix \citep{nguyen-etal-2018-tsix} & 26 days &30 &925 &36  &Extractive &109 \\
    \end{tabular}}
    \caption{Overview of other Twitter datasets.}
    \label{Overview}
\end{table*}
We next compare our corpus against the most recent and popular Twitter datasets for summarisation in \textit{Table} \ref{Overview}. To the best of our knowledge there are currently no abstractive summarisation Twitter datasets for either event or opinion summarisation. While we primarily focussed on the collection of opinionated clusters, some of the clusters we had automatically identified as opinionated were not deemed to be so by our annotators. Including the non-opinionated clusters helps expand the depth and range of Twitter datasets for summarisation.

Compared to the summarisation of product reviews and news articles, which has gained recognition in recent years because of the availability of large-scale datasets and supervised neural architectures, Twitter summarisation remains a mostly uncharted domain with very few datasets curated. \citet{inouye}\footnote{It is unclear whether the full corpus is available: our statistics were calculated based on a sample of 100 posts for each topic, but the original paper mentions that 1500 posts for each topic were initially collected.} collected the tweets for the top ten trending topics on Twitter for 5 days and manually clustered these. The SMERP dataset \citep{smerp-ecir-2017} focuses on topics on post-disaster relief operations for the 2016 earthquakes in central Italy. Finally, TSix \citep{nguyen-etal-2018-tsix} is the dataset most similar to our work as it covers, but on a smaller scale, several popular topics that are deemed relevant to news providers.

Other Twitter summarisation datasets include: \citep{zubiaga2012realtime, football} on summarisation of football matches, \citep{olariu-2014-efficient} on real-time summarisation for Twitter streams. These datasets are either publicly unavailable or unsuitable for our summarisation task\footnote{Comparing to live stream summarisation where millions of posts are used as input, we focus on summarisation of clusters of maximum 50 posts.}. 

\noindent\textbf{Summary type}. These datasets exclusively contain extractive summaries, where several tweets are chosen as representative per cluster. This results in summaries which are often verbose, redundant and information-deficient. As shown in other domains \citep{grusky-etal-2018-newsroom, narayan-etal-2018-dont}, this may lead to bias towards extractive summarisation techniques and hinder progress for abstractive models. Our corpus on COVID-19 and Election data aims to bridge this gap and introduces an abstractive gold standard generated by journalists experienced in sub-editing. 

\noindent\textbf{Size}. The average number of posts in our clusters is 30, which is similar to the TSix dataset and in line with the empirical findings by \citet{inouye}, who recommend 25 tweets/cluster. Having clusters with a much larger number of tweets makes it harder to apply our guidelines for human summarisation. 
To the best of our knowledge, our combined corpus (COVID-19 \& Election) is currently the biggest human generated corpus for microblog summarisation.

\noindent\textbf{Time-span}. Both COVID-19 and Election partitions were collected across year-long time spans. This is in contrast to other datasets, which have been constructed in brief time windows, ranging from 3 days to a month. This emphasises the longitudinal aspect of the dataset, which also allows topic diversity as 153 keywords and accounts were tracked through time.

\normalsize
\section{Defining Model Baselines}
As we introduce a novel summarisation task (MOS), the baselines featured in our experiments are selected from domains tangential to microblog opinion summarisation, such as news articles, Twitter posts and product reviews (See Sec.\ref{Sec_2}). In addition, the selected models represent diverse summarisation strategies: abstractive or extractive, supervised or unsupervised, multi-document (MDS) or single-document summarisation (SDS). Note that most SDS models enforce a length limit (1024 characters) over the input which makes it impossible to summarise the whole cluster of tweets. We address this issue by only considering the most relevant tweets ordered by topic relevance. The latter is computed using the Kullback-Leibler divergence with respect to the topical word distribution of the cluster in the GSDMM-LDA clustering algorithm \citep{gsdmmlda:1}.

The summaries were generated such that their length matches the average length of the gold standard. Some model parameters (such as Lexrank) only allow sentence-level truncation, in which case the length matches the average number of sentences in the gold standard. For models that allow a word limit to the text to be generated (BART, Pegasus, T5), a minimum and maximum number of tokens was imposed such that the generated summary would be within [90$\%$, 110$\%$] of the gold standard length.


\subsection{Heuristic Baselines}
\textbf{Extractive Oracle}: this baseline uses the gold summaries to extract the highest scoring sentences from a cluster of tweets. We follow~\citet{zhong-etal-2020-extractive} and rank each sentence by its average ROUGE-\{1,2,L\} recall scores. We then consider the highest ranking 5 sentences to form combinations of k\footnote{For opinionated clusters, we set k=3 and for non-opinionated k=1.} sentences, which are re-evaluated against the gold summaries. $k$ is chosen to equal the average number of sentences in the gold standard. The highest scoring summary with respect to the average ROUGE-\{1,2,L\} recall scores is assigned as the oracle.

\noindent\textbf{Random}: $k$ sentences are extracted at random from a tweet cluster. We report the mean result over 5 iterations with different random seeds.

\subsection{Extractive Baselines}
\textbf{LexRank} \citep{lexrank} constructs a weighed connectivity graph based on cosine similarities between sentence TF-IDF representations. 

\noindent\textbf{Hybrid TF-IDF} \citep{inouye} is an unsupervised  model designed for Twitter
, where a post is summarised as the weighted mean of its TF-IDF word vectors.

\noindent\textbf{BERTSumExt} \citep{liu-lapata-2019-text} is an SDS model comprising a BERT \citep{Devlin2019BERTPO}-based encoder stacked with Transformer layers to capture document-level features for sentence extraction. We use the model trained on CNN/Daily Mail \citep{cnndm}.

\noindent\textbf{HeterDocSumGraph} \citep{wang-etal-2020-heterogeneous} introduces the heterogenous graph neural network, which is constructed and iteratively updated using both sentence nodes and nodes representing other semantic units, such as words.
We use the MDS model trained on Multi-News \citep{fabbri-etal-2019-multi}. 

\noindent \textbf{Quantized Transformer} \citep{angelidis-etal-2021-extractive} combines Transformers \citep{transformers} and Vector-Quantized Variational Autoencoders for the summarisation of popular opinions in reviews. We trained QT on the MOS corpus.

\subsection{Abstractive Baselines}
\textbf{Opinosis} \citep{ganesan-etal-2010-opinosis} is an unsupervised MDS model. 
Its graph-based algorithm identifies valid paths in a word graph and returns the highest scoring path with respect to redundancy.  

\noindent\textbf{PG-MMR}  \citep{lebanoff-etal-2018-adapting} adapts the single document setting for multi-documents by introducing `mega-documents' resulting from concatenating clusters of texts. The model combines an abstractive SDS pointer-generator network with an MMR-based extractive component.

\noindent\textbf{PEGASUS} \citep{Zhang2020PEGASUSPW} introduces gap-sentences as a pre-training objective for summarisation. 
It is then fine-tuned for 12 downstream summarisation domains.
We chose the model pre-trained on Reddit TIFU \citep{Kim:2019:NAACL-HLT}. 

\noindent\textbf{T5} \citep{raffel2020exploring} adopts a unified approach for transfer learning on language-understanding tasks. For summarisation, the model is pre-trained on the  Colossal Clean Crawled Corpus \citep{raffel2020exploring} and then fine-tuned on CNN/Daily Mail.

\noindent\textbf{BART} \citep{lewis-etal-2020-bart} is pre-trained on several evaluation tasks, including summarisation. 
With a bidirectional encoder and GPT2, BART is considered a generalisation of BERT. We use the BART model pre-trained on CNN/Daily Mail.

\noindent\textbf{SummPip} \citep{summpip} is an MDS unsupervised model which constructs a sentence graph following Approximate Discourse Graph and deep embedding methods.
After spectral clustering of the sentence graph, summary sentences are generated through a compression step of each cluster of sentences. 

\noindent\textbf{Copycat} \citep{brazinskas-etal-2020-unsupervised} is a Variational Autoencoder model trained in an unsupervised setting to capture the consensus opinion in product reviews for Yelp and Amazon. We train it on the MOS corpus.

\section{Evaluation Methodology}
Similar to other summarisation work \citep{fabbri-etal-2019-multi, grusky-etal-2018-newsroom}, we perform both automatic and human evaluation of models. 
Automatic evaluation is conducted on a set of 200 clusters: each partition of the test (COVID-19 Opinionated, COVID-19 Non-opinionated, Election Opinionated, Election Non-opinionated) contains 50 clusters uniformly sampled from the total corpus. For the human evaluation, only the 100 opinionated clusters are evaluated.

\subsection{Automatic Evaluation}
Word overlap is evaluated according to the harmonic mean $F_1$ of ROUGE-1, 2, L\footnote{We use ROUGE-1.5.5 via the \textit{pyrouge} package: \url{https://github.com/bheinzerling/pyrouge}} \cite{lin-2004-rouge} as reported elsewhere \citep{narayan-etal-2018-dont,gholipour-ghalandari-etal-2020-large,Zhang2020PEGASUSPW}. Work by \citet{tay-etal-2019-red} acknowledges the intractability of ROUGE in opinion text summarisation as sentiment-rich language uses a vast vocabulary that does not rely on word matching. This issue is mitigated by \citep{booksum, bhandari-etal-2020-evaluating}, who use semantic similarity as an additional assessment of candidate summaries. Similarly, we use text generation metrics BLEURT \citep{sellam-etal-2020-bleurt} and BERTScore\footnote{BERTScore has a narrow score range, which makes its interpretation more difficult than for BLEURT. Since both metrics produce similar rankings, BERTScore can be found in Appendix \ref{Appen_B}.} \citep{bert-score} to assess semantic similarity.

\subsection{Human Evaluation}
Human evaluation is conducted to assess the quality of summaries with respect to three objectives: 1) linguistic quality, 2) informativeness and 3) ability to identify opinions.  We conducted two human evaluation experiments: the first (\ref{non_FT_guidelines}) assesses the gold standard and non-fine-tuned model summaries on a rating scale, while the second (\ref{FT_model_guidelines}) addresses the advantages and disadvantages of fine-tuned model summaries via Best-Worst Scaling. Four and three experts 
were employed for the two experiments, respectively.

\subsubsection{Evaluation of Gold Standard \& Models}
\label{non_FT_guidelines}
The first experiment focused on assessing the gold standard and best models from each summarisation type: Gold, LexRank (best extractive), SummPip (best unsupervised abstractive) and BART (best supervised).

\vspace{.1cm}
\noindent\textbf{Linguistic quality} measures 4 syntactic dimensions, which were inspired by previous work on summary evaluation. Similar to DUC \citep{DUC}, each summary was evaluated with respect to each criterion below on a 5-point scale.
\vspace{-0.3cm}
\begin{itemize}
    \item \textit{Fluency} \citep{grusky-etal-2018-newsroom}: Sentences in the summary ``should have no formatting problems, capitalization errors or obviously ungrammatical sentences (e.g., fragments, missing components) that make the text difficult to read.''
    \vspace{-0.3cm}
    \item \textit{Sentential Coherence} \citep{grusky-etal-2018-newsroom}: A sententially coherent summary should be well-structured and well-organized. The summary should not just be a heap of related information, but should build from sentence to sentence to a coherent body of information about a topic.
 \vspace{-0.3cm}
    \item \textit{Non-redundancy} \citep{DUC}: A non-redundant summary should contain no duplication, i.e., there should be no overlap of information between its sentences.
 \vspace{-0.3cm}
    \item \textit{Referential Clarity} \citep{DUC}: It should be easy to identify who or what the pronouns and noun phrases in the summary are referring to. If a person or other entity is mentioned, it should be clear what their role is in the story.
\end{itemize}

\vspace{.1cm}
\noindent\textbf{Informativeness} is defined as the amount of factual information displayed by a summary. To measure this, we use a Question-Answer algorithm \citep{questiongeneration20} as a proxy. Pairs of questions and corresponding answers are generated from the information nuggets of each cluster. Since we want to assess whether the summary contains factual information, only information nuggets belonging to the `WHAT', `WHO', `WHERE' are selected as input. We chose not to include the entire cluster as input for the QA algorithm, as this might lead the informativeness evaluation to prioritize irrelevant details in the summary. Each cluster in the test set is assigned a question-answer pair and each system is then scored based on the percentage of times its generated summaries contain the answer to the question. Similar to factual consistency \citep{wang-etal-2020-asking}, informativeness penalises incorrect answers (hallucinations), as well as the lack of a correct answer in a summary.

\vspace{.1cm}
\noindent As \textbf{Opinion} is a central component for our task, we want to assess the extent to which summaries contain opinions. Assessors report whether summaries identify any majority or minority opinions\footnote{Note that whether the identified minority or majority opinions are correct is not evaluated here. This is done in Section \ref{FT_model_guidelines}.}. A summary contains a majority opinion if most of its sentences express this opinion or if it contains specific terminology (`The majority is/ Most users think...' etc.), which is usually learned during the fine-tuning process. Similarly, a summary contains a minority opinion if at least one of its sentences expresses this opinion or it contains specific terminology (`A minority/ A few users' etc.). The final scores for each system are the percentage of times the summaries contain majority or minority opinions, respectively. 

\subsubsection{Best-Worst Evaluation of Fine-tuned Models}
\label{FT_model_guidelines}
The second human evaluation assesses the effects of fine-tuning on the best supervised model, BART. The experiments use non-fine-tuned BART (BART), BART fine-tuned on $10\%$ of the corpus (BART FT10$\%$) and BART fine-tuned on $70\%$ of the corpus (BART FT70$\%$).

As all the above are versions of the same neural model, Best-Worst scaling is chosen to detect subtle improvements, which cannot otherwise be quantified as reliably by traditional ranking scales \citep{kiritchenko-mohammad-2017-best}. An evaluator is shown a tuple of 3 summaries (BART, BART FT70$\%$, BART FT30$\%$) and asked to choose the best/worst with respect to each criteria. To avoid any bias, the summary order is randomized for each document following~\citet{van-der-lee-etal-2019-best}. The final score is calculated as the percentage of times a model is scored as the best, minus the percentage of times it was selected as the worst \citep{Orme2009MaxDiffA}. In this setting, a score of 1 represents the unanimously best, while -1 is unanimously the worst.

The same criteria as before are used for \textbf{linguistic quality} and one new criterion is added to assess \textbf{Opinion}. We define \textit{Meaning Preservation} as the extent to which opinions identified in the candidate summaries match the ones identified in the gold standard. We draw a parallel between the \textit{Faithfulness} measure \citep{maynez-etal-2020-faithfulness}, which assesses the level of hallucinated information present in summaries and \textit{Meaning Preservation}, which assesses the extent of hallucinated opinions.

\section{Results}
\label{Sec_6}

\subsection{Automatic Evaluation}
Results for the automatic evaluation are shown in 
\textit{Table}~\ref{tab:rouge_bleurt-scores}. 

\begin{table*}[tb]
    \centering
    \resizebox{\linewidth}{!}{%
    \scriptsize
    \begin{tabular}{lcccccccccccccccc} 
    \toprule
    & \multicolumn{4}{c}{\textbf{COVID-19 Opinionated (CO)}} & \multicolumn{4}{c}{\textbf{COVID-19 Non-opinionated (CNO)}} & \multicolumn{4}{c}{\textbf{Election Opinionated (EO)}} & \multicolumn{4}{c}{\textbf{Election Non-opinionated (ENO)}} \\
    \cmidrule(lr){2-5} 
    \cmidrule(lr){6-9}
    \cmidrule(lr){10-13}
    \cmidrule(lr){14-17}
    \textbf{Models} & $\textrm{R-1}_{f_1}$ & $\textrm{R-2}_{f_1}$ & $\textrm{R-L}_{f_1}$ & $\textrm{BLEURT}$ & $\textrm{R-1}_{f_1}$ & $\textrm{R-2}_{f_1}$ & $\textrm{R-L}_{f_1}$ & $\textrm{BLEURT}$& $\textrm{R-1}_{f_1}$ & $\textrm{R-2}_{f_1}$ & $\textrm{R-L}_{f_1}$ & $\textrm{BLEURT}$& $\textrm{R-1}_{f_1}$ & $\textrm{R-2}_{f_1}$ & $\textrm{R-L}_{f_1}$ & $\textrm{BLEURT}$\\
    \midrule
    \multicolumn{17}{c}{Heuristics} \\
    \midrule
    
    Gold (195 char)& & & & & & & & & & & & &&&& \\
    Random Sentences (204 char)&13.55 & 1.09 &9.22& -.660 &7.30 &0.70 &5.97&-.968 &11.82 &0.80 &8.27&-.576 &6.75 &0.89 &5.69 &-.592 \\
    Extractive Oracle (289 char)& 15.45 & 1.67 & 10.29&-.382 & 11.80 & 1.38 & 9.27&-.510 & 15.33 & 1.60 & 10.12&\textbf{-.146}& 10.06 & \textbf{2.15} & 8.46 &\textbf{-.056}\\
    \midrule
    \multicolumn{17}{c}{Extractive Unsupervised Models} \\
    \midrule

     LexRank (265 char)&\cellcolor{Gray}16.41 & \cellcolor{Gray}1.48 & \cellcolor{Gray}10.89&\cellcolor{Gray}-.560 &\cellcolor{Gray} 10.87 & 1.01 & \cellcolor{Gray}8.76&-.849 &14.27 &\cellcolor{Gray}1.15 &\cellcolor{Gray}9.62&-.418 &\cellcolor{Gray}9.11 &1.08 &\cellcolor{Gray}7.41 &-.456\\
   
    Hybrid TF-IDF (277 char)&12.87 &1.26 &8.85&-.608 &9.33 &0.83 &7.51&-.745 &12.06 &1.12 &8.42&-.430 &7.93 &\cellcolor{Gray}1.13 &6.56&\cellcolor{Gray}-.298 \\
     Quantized Transformer (273 char) &  14.23 &  1.03 &  9.55&-.621 &  9.85 & 0.96 &  7.83&-.857 &  14.78 &  1.08 &  9.45&-.468 &  8.69 &0.81 &6.79&-.668\\
    \midrule
    \multicolumn{17}{c}{Extractive Supervised Models}\\
    \midrule

    BERTSumExt (225 char)& 14.22 &1.31 & 9.68&-.571 &9.78 &\cellcolor{Gray}1.11 &7.70&\cellcolor{Gray}-.699 &11.93 &1.10 &8.47&\cellcolor{Gray}-.384 &8.06 &1.00 &6.63 &-.407\\
    HeterDocSumGraph (295 char)&15.13 &1.19 &9.79& -.748&10.05 &0.88 &7.79&-.867 &\cellcolor{Gray}14.28 &0.96 &9.15&-.564 &8.40 &0.72 &6.86 &-.626\\
      
    \midrule
    \multicolumn{17}{c}{Abstractive Unsupervised Models} \\
    \midrule
     Opinosis (215 char)&12.45 &1.14 &8.86&-.534 &8.35 &0.73 &6.99&-.673 &11.34 &1.00 &8.15&-.537 &6.69 &0.95 &5.66&-.518 \\
     SummPip (236 char)&12.96 &1.37 &9.32&-.488 &\cellcolor{Gray}11.30 &1.46 &\cellcolor{Gray}9.09&\cellcolor{Gray}-.559 &13.05 &1.15 &8.90&-.409 &\cellcolor{Gray}9.93 &\cellcolor{Gray}1.36 &\cellcolor{Gray}7.74 &\cellcolor{Gray}-.228\\
    
    Copycat (153 char) &  12.47&  1.31 &  9.41&-.552 & 10.99 &  1.32 &  9.25&-.621 &  14.05 & 1.56 &  10.25&-.503 &  7.48 &1.10 &6.36&-.316\\
    
     \midrule
    \multicolumn{17}{c}{Abstractive Supervised Models} \\
    \midrule
    PG-MMR (238 char) &11.93 &1.08 &8.93&\cellcolor{Gray}-.450 &9.68 &1.37 &8.01&-.578 &12.36 &1.07 &8.73&\cellcolor{Gray}-.400 &8.14 &1.04 &6.86 &-.302\\
    Pegasus (216 char)&13.78 &1.40 &9.78&-.535 &10.37 &1.41 &8.61&-.616 &12.68 &\cellcolor{Gray}1.23 &\cellcolor{Gray}9.28&-.481 &9.12 &1.11 &7.34 &-.283\\
    T5 (206 char)&14.25 &1.31 &9.97&-.530 &9.11 &1.21 &7.72&-.669 &12.99 &1.06 &8.82& -.470&8.59 &1.15 &7.06 &-.347\\
    BART (237 char)&\cellcolor{Gray}15.95 & \cellcolor{Gray}1.46 & \cellcolor{Gray}10.74&-.521 & 10.41 & \cellcolor{Gray}1.55 & 8.48&-.576 &\cellcolor{Gray}13.71 &1.18 &9.09&-.409 &9.11 &1.15 &7.37 &-.372\\
    
    \midrule
    \multicolumn{17}{c}{Fine-tuned Models} \\
    \midrule
    BART (FT 10$\%$) (245 char)&\cellcolor{Gray}21.53 & \cellcolor{Gray}\textbf{3.86} & \cellcolor{Gray}\textbf{14.76}&\cellcolor{Gray}\textbf{-.257} & \cellcolor{Gray}\textbf{15.49} & \cellcolor{Gray}\textbf{2.61} & 12.04&-.449 &19.77 &2.99 &13.11&-.209 &12.31 &\cellcolor{Gray}\textbf{1.87} &\cellcolor{Gray}\textbf{9.62}&-.081\\
    
    BART (FT 70$\%$) (246 char)&\cellcolor{Gray}\textbf{21.54} & 3.74 & 14.54&-.259 & 15.31 & 2.54 & \cellcolor{Gray}\textbf{12.09}&\cellcolor{Gray}\textbf{-.439} &\cellcolor{Gray}\textbf{20.59} &\cellcolor{Gray}\textbf{3.42} &\cellcolor{Gray}\textbf{13.63}&\cellcolor{Gray}\textbf{-.183} &\cellcolor{Gray}\textbf{12.37} &1.72 &9.58 &\cellcolor{Gray}\textbf{-.071}\\
    \bottomrule
    \end{tabular}
    }%
    \caption{
    Performance on the \textbf{test set} of baseline models evaluated with automatic metrics: ROUGE-n (R-n) and BLEURT. The best model \hl{from each category} (Extractive, Abstractive, Fine-tuned) and \hl{\textbf{overall}} are highlighted.
    }
    \label{tab:rouge_bleurt-scores}
\end{table*}

\begin{table}[tb]
    \centering
    \resizebox{\linewidth}{!}{%
    \scriptsize
    \begin{tabular}{lcccccccc} 
    \toprule
    & \multicolumn{4}{c}{\textbf{COVID-19 Opinionated (CO)}} &  \multicolumn{4}{c}{\textbf{Election Opinionated (EO)}} \\
    \cmidrule(lr){2-5} 
    \cmidrule(lr){6-9}
    \textbf{Models} & $\textrm{R-1}_{f_1}$ & $\textrm{R-2}_{f_1}$ & $\textrm{R-L}_{f_1}$ & $\textrm{BLEURT}$ & $\textrm{R-1}_{f_1}$ & $\textrm{R-2}_{f_1}$ & $\textrm{R-L}_{f_1}$ & $\textrm{BLEURT}$\\
\midrule
    &\multicolumn{8}{c}{Main Story} \\
    \midrule
     BART (FT 10$\%$) &\textbf{11.43} &\textbf{2.49} &\textbf{9.95} &\textbf{-.082} &  \textbf{9.82} &  \textbf{1.72} &  \textbf{8.31}& -.185\\
      BART (FT 70$\%$) &11.18 &2.29 &9.57 &-.137 & 9.55 &  \textbf{1.70} &  8.19& \textbf{-.104}\\
    \midrule
    &\multicolumn{8}{c}{Majority Opinion} \\
    \midrule
     BART (FT 10$\%$) &\textbf{20.25} &\textbf{4.28} &\textbf{16.86} &\textbf{-.487} &  17.88 &  3.11 &  14.57& -.442\\
     BART (FT 70$\%$) 19.74 4.06 &16.18 &-.505 & \textbf{19.13} &  \textbf{3.74} &  \textbf{15.60}& \textbf{-.392}\\

    \midrule
    &\multicolumn{8}{c}{Minority Opinion(s)} \\
    \midrule
   BART (FT 10$\%$) &\textbf{19.05} &4.66 &\textbf{15.87} &\textbf{-.544} & 15.26 & 3.97 &  13.34& -.791\\
    BART (FT 70$\%$) &18.70  &\textbf{4.81}  &\textbf{15.83} &-.643 & \textbf{15.98} &  \textbf{4.63} &  \textbf{14.01}& \textbf{-.604}\\
 
    \bottomrule
    \end{tabular}
    }%
    \caption{
    Performance of fine-tuned models per each summary component (Main Story, Majority Opinion, Minority Opinion(s)) on the \textbf{test set} evaluated with automatic metrics: ROUGE-n (R-n) and BLEURT.
    }
    \label{tab:rouge_bleurt-scores_component}
\end{table}
\noindent\textbf{Fine-tuned Models} Unsurprisingly, the best performing models are ones that have been fine-tuned on our corpus: \textit{BART (FT70$\%$)} and \textit{BART (FT10$\%$)}. Fine-tuning has been shown to yield competitive results for many domains \citep{booksum, fabbri-etal-2021-improving}, including ours. In addition, one can see that only the fine-tuned abstractive models are capable of outperforming the \textit{Extractive Oracle}, which is set as the upper threshold for extractive methods. Note that on average, the Oracle outperforms the Random summariser by a $59\%$ margin, which only fine-tuned models are able to improve on, with $112\%$ for \textit{BART (FT10$\%$)} and $114\%$ for \textit{BART (FT70$\%$)}.
We hypothesise that our gold summaries' template format poses difficulties for off-the-shelf models and fine-tuning even on a limited portion of the corpus produces summaries that follow the correct structure (See Table \ref{Summary_example1} and Appendix C for examples). We include comparisons between the performance of \textit{BART (FT10$\%$)} and \textit{BART (FT70$\%$)} on the individual components of the summary in Table \ref{tab:rouge_bleurt-scores_component}. \footnote{We do not include other models in the summary component-wise evaluation because it is impossible to identify the Main Story, Majority Opinion and Minority Opinions in non-fine-tuned models.}

\noindent\textbf{Non-Fine-tuned Models} Of these, \textit{SummPip} performs the best across most metrics and datasets with an increase of $37\%$ in performance over the random model, followed by \textit{LexRank} with an increase of $29\%$. Both models are designed for the multi-document setting and benefit from the common strategy of mapping each sentence in a tweet from the cluster into a node of a sentence graph. However, not all graph mappings prove to be useful: summaries produced by \textit{Opinosis} and \textit{HeterDocSumGraph}, which employ a word-to-node mapping, do not correlate well with the gold standard. The difference between word and sentence-level approaches can be partially attributed to the high amount of spelling variation in tweets, making the former less reliable than the latter.

\noindent\textbf{ROUGE vs BLEURT} The performance on \texttt{ROUGE} and \texttt{BLEURT} is tightly linked to the data differences between COVID-19 and Election partitions of the corpus. Most models achieve higher \texttt{ROUGE} scores and lower \texttt{BLEURT }scores on the COVID-19 than on the Election dataset. An inspection of the data differences reveals that COVID-19 tweets are much longer than Election ones (169 \textit{vs} 107 char), as the latter had been collected before the increase in length limit from 140 to 280 char in Twitter posts. This is in line with findings by \citet{sun-etal-2019-compare}, who revealed that high \texttt{ROUGE} scores are mostly the result of longer summaries rather than better quality summaries.

\subsection{Human Evaluation}

\begin{table*}[!h]
    \centering
    \resizebox{.9\textwidth}{!}{%
    \begin{tabular}{cccccccc}
\hline
        \textbf{Model} & \textbf{Fluency} & \textbf{Sentential Coherence} & \textbf{Non-redundancy} & \textbf{Referential Clarity} & \textbf{Informativeness} & \textbf{Majority} & \textbf{Minority}\\
        \hline
         Gold &4.52 &4.63&4.85&4.31&57$\%$& 86$\%$ & 64$\%$\\
         \hline
         Lexrank &3.03&2.43&3.10&2.55&58$\%$& 15$\%$& \textbf{62}$\%$\\
         BART &\textbf{3.24}&\textbf{2.76}&\textbf{3.46}&3.01&67$\%$& 8$\%$ &60$\%$\\
         SummPip &2.73 &2.70 &2.53 &\textbf{3.37}& \textbf{69}$\%$& \textbf{32}$\%$& 36$\%$\\
         \hline
    \end{tabular}}
    \caption{Evaluation of Gold Standard and Models: Results}
    \label{tab:non_FT_models}
\end{table*}
\normalsize

\begin{table*}[!h]
    \centering
    \scriptsize
    \begin{tabular}{c|c|c|c|c|c}
         \hline
         Model & Fluency & Sentential Coherence & Non-redundancy & Referential Clarity & Meaning Preservation \\
         \hline
         BART& -0.76&-0.65 &\textbf{0.15} &-0.42& -0.54 \\
         BART FT 10$\%$&0.30 &0.22 &-0.11 &\textbf{0.25} & 0.14\\
         BART FT 70$\%$ &\textbf{0.44} &\textbf{0.43} &-0.04 &0.17 &\textbf{0.40} \\
         \hline
    \end{tabular}
    \caption{Best-Worst Evaluation of Fine-tuned models: Results}
    \label{tab:FT_models}
\end{table*}
\normalsize
\noindent\textbf{Evaluation of Gold Standard \& Models} 
Table~\ref{tab:non_FT_models} shows the comparison between the gold standard and the best performing models against a set of criteria (See \ref{non_FT_guidelines}). As expected, the human-authored summaries (Gold) achieve the highest scores with respect to all linguistic quality and structure-based criteria. However, the gold standard fails to capture informativeness as well as its automatic counterparts, which are, on average, longer and thus may include more information. Since \textit{BART} is previously pre-trained on CNN/DM dataset of news articles, its output summaries are more fluent, sententially coherent and contain less duplication than the unsupervised models \textit{Lexrank} and \textit{SummPip}. We hypothesise that \textit{SummPip} achieves high referential clarity and majority scores as a trade-off for its very low non-redundancy (high redundancy).

\noindent\textbf{Best-Worst Evaluation of Fine-tuned Models} The results for our second human evaluation are shown in Table \ref{tab:FT_models} using the guidelines presented in \ref{FT_model_guidelines}. The model fine-tuned on more data \textit{BART (FT70$\%$)} achieves the highest fluency and sentential coherence scores. As seen in Table \ref{Summary_example1}, the summary produced by \textit{BART (FT70$\%$)} contains complete and fluent sentences, unlike its counterparts. Most importantly, fine-tuning yields better alignment with the gold standard with respect to meaning preservation, as the fine-tuned models \textit{BART (FT70$\%$)} and \textit{BART (FT10$\%$)} learn how to correctly identify and summarise the main story and the relevant opinions in a cluster of tweets. In the specific example, non-fine-tuned \textit{BART} introduces a lot of irrelevant information (`industrial air pollution',`google, apple rolling out covid') to the main story and offers no insight into the opinions found in the cluster of tweets, whereas both fine-tuned models correctly introduce the Main Story and both partially identify the Majority Opinion (`great idea' for anti-maskers `to dig graves'). However, we note that the fine-tuning process does not lead to increased performance with respect to all criteria; non-redundancy is compromised and referential clarity stops improving after a certain amount of training data. As observed in the example, \textit{BART (FT70$\%$)} contains duplicated content `think this is a great idea. What a great idea!'. Wilcoxon signed rank tests with $p<0.05$ and $p<0.10$ are used for significance testing between all pairs of models. We note that most pairwise differences are significant at $p<0.05$, while \textit{BART (FT70$\%$)} and BART NFT differences are significant at $p<0.10$ for non-redundancy. The only two exceptions are referential clarity and non-redundancy between \textit{BART (FT70$\%$)} and \textit{BART (FT10$\%$)} where both fine-tuned models perform similarly.

\begin{table}[!h]
    \tiny
    \begin{tabular}{p{7.5cm}}
        \toprule
        \multirow{3}{7.5cm}{\textbf{Human Summary}: Anti-maskers are forced to dig graves for Covid19 victims in Indonesia. The majority of Twitter users think it is a good idea that will cause a reality check and make people rethink their ideas.}
        \\ \\ \\
        \midrule
        \multirow{4}{7.5cm}{\textbf{BART FT $70\%$}: Anti-maskers in Indonesia are forced to dig graves for Covid-19 victims as punishment for not wearing mask. The majority think this is a great idea. What a great idea! It's a good way to get people to reconsider their misinformed opinions. We should do this in the US.}
        \\ \\ \\ \\ 
        \multirow{4}{7.5cm}{\textbf{BART FT $10\%$}: Anti-maskers forced to dig graves for Covid-19 victims in Indonesia as punishment for refusing to wear a mask. The majority believe that this is a good idea, and that it will hopefully bring about an end to the need for masks. A minority do not believe it is necessary to wear a}
        \\ \\ \\ \\ 
        \multirow{4}{7.5cm}{\textbf{BART}: Covid-19 can be more deadly for people living in areas with industrial air pollution. Anyone refusing to wear a mask is made to dig graves for covid-19 victims as punishment in east java. as domestic violence spikes, many victims and their children have nowhere to live. google, apple rolling out covid-}
        \\ \\ \\ \\
        \bottomrule
    \end{tabular}
    \caption{BART Summary Examples for the same cluster of tweets.}
    \label{Summary_example1}
\end{table}
\normalsize

\section{Error Analysis}
Error analysis is carried out on 30 fine-tuned BART summaries from a set of 15 randomly sampled clusters. The results are found in Table \ref{Summary_example2}.

\begin{table}[!h]
    \tiny
    \begin{tabular}{p{1cm}p{0.3cm}p{5.4cm}}
        \toprule
        Error type & Freq. & Example \\
        \midrule
        Intrinsic &4/30&\textbf{\ul{Example 1}}\\
        Hallucination&&\multirow{6}{5.4cm}{\textbf{Generated Summary}: United States surpasses six million coronavirus cases and deaths and remains at the top of the global list of countries with the most cases and deaths \ul{The majority are pleased} to see the US still leads the world in terms of cases and deaths, with 180,000 people succumbing to Covid-19.}
        \\ \\ \\ \\ \\ \\
        \midrule
        Extrinsic &4/30&\textbf{\ul{Example 2}}\\
        Hallucination&&\multirow{5}{5.4cm}{\textbf{Generated Summary}: Sex offender Rolf Harris is involved in a prison brawl after absconding from open jail. The majority think Rolf Harris deserves to be spat at and called a "nonce" and \ul{a "terrorist" for absconding from open prison}. A minority are putting pressure on }
        \\ \\ \\ \\ \\
        \midrule
        Information&12/30&\textbf{\ul{Example 3}}\\
        Loss&&\multirow{4}{5.4cm}{\textbf{Human Summary}: Miley Cyrus invited a homeless man on stage to accept her award. Most people thought it was a lovely thing to do and it was emotional. \ul{A minority think that it was a publicity stunt.}}
        \\ \\ \\ \\
        &&\multirow{6}{5.4cm}{\textbf{Generated Summary}: Miley Cyrus had homeless man accept Video of the Year award at the MTV Video Music Awards. The majority think it was fair play for Miley Cyrus to allow the homeless man to accept the award on her behalf. She was emotional and selfless. The boy band singer cried and thanked him for accepting the}
        
        \\ \\ \\ \\ \\ 
     \bottomrule
    \end{tabular}
    \caption{Error Analysis: Frequency of errors and representative summary examples for each error type.}
    \label{Summary_example2}
\end{table}

\normalsize

\noindent\textbf{Hallucination} Fine-tuning on the MOS corpus introduces hallucinated content in 8 out of 30 manually evaluated summaries. Generated summaries contain opinions that prove to be either false or unfounded after careful inspection of the cluster of tweets. We follow the work of \citet{maynez-etal-2020-faithfulness} in classifying hallucinations as either intrinsic (incorrect synthesis of information in the source) or extrinsic (external information not found in the source). Example 1 in Table \ref{Summary_example2} is an instance of an intrinsic hallucination: the majority opinion is wrongly described as `pleased', despite containing the correct facts regarding US coronavirus cases. Next, Example 2 shows that Rolf Harris `is called a terrorist', which is confirmed to be an extrinsic hallucination as none of the tweets in the source cluster contain this information.

\noindent\textbf{Information loss} Information loss is the most frequent error type. As outlined in~\citep{booksum}, the majority of current summarisation models face length limitations (usually 1024 characters) which are detrimental for long-input documents and tasks. Since our task involves the detection of all opinions within the cluster, this weakness may lead to incomplete and less informative summaries as illustrated in Example 3 from Table \ref{Summary_example2}. The candidate summary does not contain the minority opinion identified by the experts in the gold standard. An inspection of the cluster of tweets reveals that most posts expressing this opinion are indeed not found in the first 1024-char allowed limit of the cluster input. 

\section{Conclusions and Future Work}
We have introduced the task of Twitter opinion summarisation and constructed the first abstractive corpus for this domain, based on template-based human summaries. Our experiments show that existing extractive models fall short on linguistic quality and informativeness while abstractive models perform better but fail to identify all relevant opinions required by the task. Fine-tuning on our corpus boosts performance as the models learn the summary structure.

In the future, we plan to take advantage of the template-based structure of our summaries to refine fine-tuning strategies. One possibility is to exploit style-specific vocabulary during the generation step of model fine-tuning to improve on capturing opinions and other aspects of interest. 

\section*{Acknowledgements}
This work was supported by a UKRI/EPSRC Turing AI Fellowship to Maria Liakata (grant no. EP/V030302/1) and The Alan Turing Institute (grant no. EP/N510129/1) through project funding and its Enrichment PhD Scheme.
We are grateful to our reviewers and action editor for reading our paper carefully and critically and thank them for their insightful comments and suggestions. We would also like to thank our annotators for their invaluable expertise in constructing the corpus and completing the evaluation tasks.

\section*{Ethics}
Ethics approval to collect and to publish extracts from social media datasets was sought and received from Warwick University Humanities \& Social Sciences Research Ethics Committee. When the corpus will be released to the research community, only tweets IDs will be made available along with associated cluster membership and summaries. Compensation rates were agreed with the annotators before the annotation process was launched. Remuneration was fairly paid on an hourly rate at the end of task.

\bibliography{tacl2018}
\bibliographystyle{acl_natbib}
\clearpage
\section*{Appendix A}
\subsection*{Summary Annotation Interface}\label{Appen_A}

\noindent\textbf{Stage 1: Reading and choosing cluster type}

\begin{figure}[!h]
\includegraphics[width=\textwidth/2]{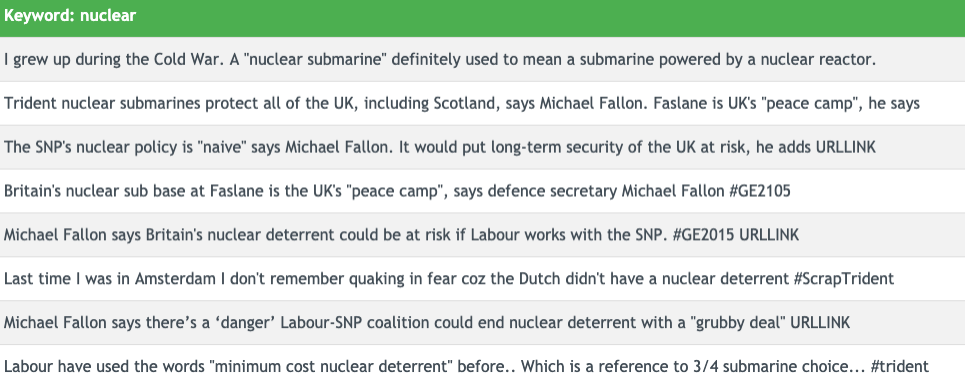}
\vspace*{-10mm}
\caption{\scriptsize Fragment of a cluster of tweets for keyword `nuclear'.}
\end{figure}

\vspace*{-5mm}

\begin{figure}[!h]
\includegraphics[width=\textwidth/2]{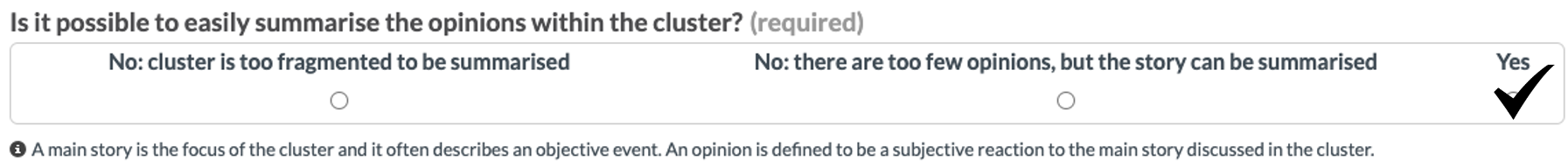}
\vspace*{-10mm}
\caption{\scriptsize Choose type of cluster `Coherent Opinionated'.}
\end{figure}

\vspace*{-3mm}
The majority of the tweets in the cluster revolve around the subject of Trident nuclear submarines. The cluster contains many opinions which can be summarised easily, hence this cluster is \textit{Coherent Opinionated}. Choose `Yes' and proceed to the next step.

\noindent\textbf{{Stage 2: Highlighting information nuggets}}
\vspace*{-2mm}
\begin{figure}[!h]
\includegraphics[width=\textwidth/2]{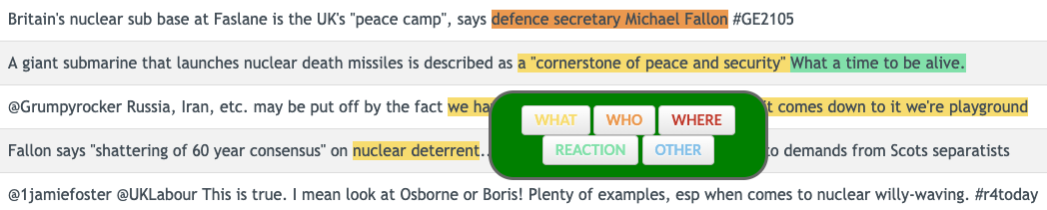}
\vspace*{-8mm}
\caption{\scriptsize Example of information nuggets: `a cornerstone of peace and security' describes the nuclear submarine (WHAT information nugget), while `defence secretary Michael Fallon' describes a person (WHO information nugget).}
\end{figure}

\vspace*{-4mm}
Highlight important information and select the relevant aspect each information nugget belongs to.

\noindent\textbf{Stage 3: Template-based Summary Writing}
\begin{figure}[!h]
\includegraphics[width=\textwidth/2]{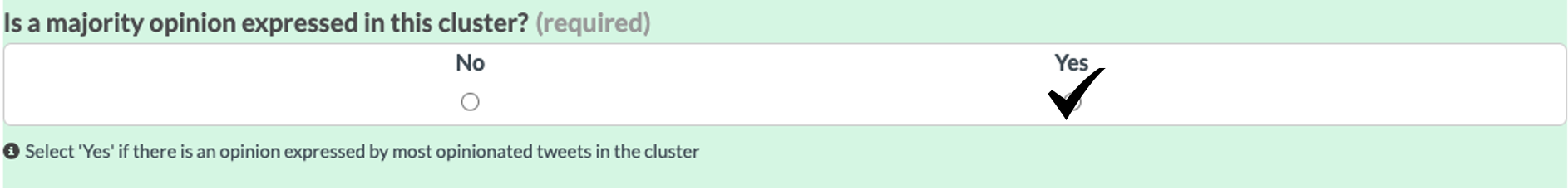}
\vspace*{-7mm}
\caption{\scriptsize Choose whether there exists a majority opinion in the cluster.}
\end{figure}
\vspace*{-4mm}
Most user reactions dismiss the Trident plan and view it as an exaggerated security measure. This forms the\textit{ Majority Opinion}. A few users express fear for UK's potential future in a nuclear war. This forms a \textit{Minority Opinion}.

\begin{figure}[!h]
\includegraphics[width=\textwidth/2]{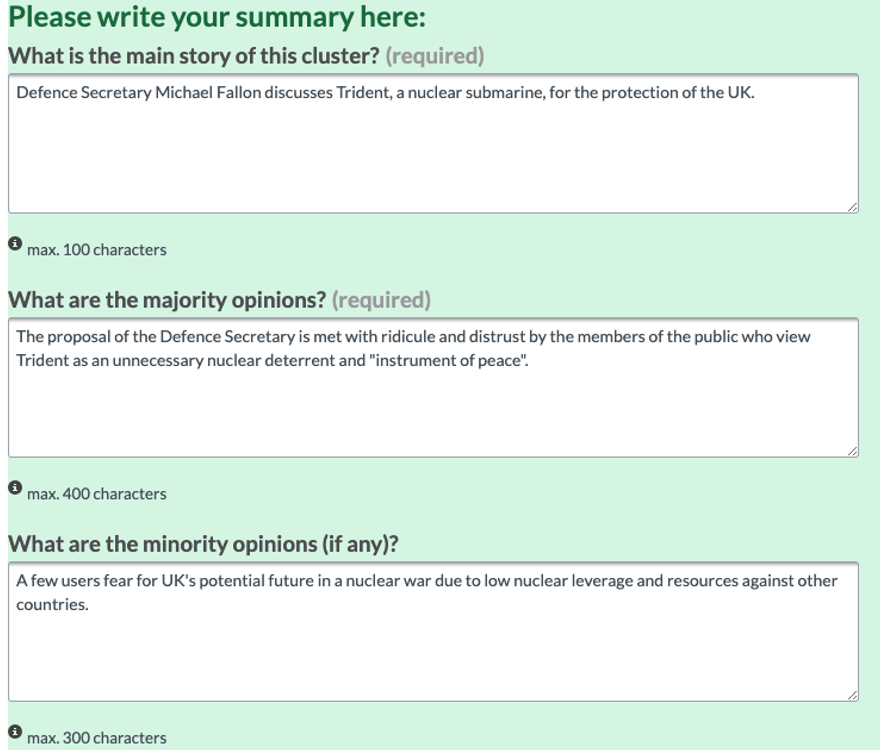}
\vspace*{-8mm}
\caption{\scriptsize Summary template of a Coherent Opinionated cluster with a majority opinion.}
\end{figure}

Write cluster summary following the structure: Main Story + Majority Opinion (+ Minority Opinions).

\section*{Appendix B}
\subsection*{Complete Results: BERTScore Evaluation}
\label{Appen_B}

\begin{table}[!h]
    \centering
    \resizebox{\linewidth}{!}{%
    \begin{tabular}{lcccc} 
    \toprule
    & \textbf{COVID-19} & \textbf{COVID-19} & \textbf{Election} & \textbf{Election} \\
    \textbf{Models} & \textbf{Opinionated} & \textbf{Non-opinionated} & \textbf{Opinionated} & \textbf{Non-opinionated}\\
    \midrule
    \multicolumn{5}{c}{Heuristics} \\
    \midrule 
    Random Sentences &0.842&0.838 &0.846 &0.861  \\
    Extractive Oracle&0.858 &0.867 &0.871 &\textbf{0.904} \\
    
    \midrule
    \multicolumn{5}{c}{Extractive Models} \\
    \midrule 
    LexRank&0.851 &0.849 &0.856 &0.868\\
    Hybrid TF-IDF &0.851 &0.853 &0.856 &0.879  \\
    BERTSumExt &0.848 &0.851 &0.859 &0.874   \\
    HeterDocSumGraph &0.839 &0.840 &0.847 &0.853 \\
    Quantized Transformer &0.840 &0.827 &0.850 & 0.856 \\
    
    \midrule
    \multicolumn{5}{c}{Abstractive Models} \\
    \midrule 
    Opinosis &0.845 &0.853 &0.846 &0.860   \\
    PG-MMR &0.853 &0.857 &0.851 &0.863   \\
    Pegasus &0.850 &0.856 &0.852 &0.869  \\
    T5 &0.850 &0.851 &0.853 &0.872 \\
    BART &0.852 &0.854 &0.856 &0.868  \\
    SummPip &0.852&0.858 &0.854 &0.878  \\
    Copycat & 0.848 & 0.852 & 0.848 & 0.872 \\
    \midrule
    \multicolumn{5}{c}{Fine-tuned Models} \\
    \midrule
    BART (FT $10\%$) & \textbf{0.873} &\textbf{0.870} &0.875 &\textbf{0.893}  \\
  
    BART (FT $70\%$) & \textbf{0.873} &\textbf{0.870} &\textbf{0.878} &0.892  \\
    
    \bottomrule
    \end{tabular}
    }%
    \caption{
    Performance on \textbf{test set} of baseline models evaluated with BERTScore.
    }
    \label{tab:bertscores}
\end{table}

\subsection*{Model Implementation Details}
T5, Pegasus and BART were implemented using the HuggingFace Transformer package \citep{wolf-etal-2020-transformers} with max sequence length of 1024 char.

Fine-tuning parameters for BART are: 8 batch size, 5 training epochs, 4 beams, enabled early stopping, 2 length penalty and no trigram repetition for the summary generation. The rest of the parameters are set as default following the configuration of \textit{ BartForConditionalGeneration}: activation function gelu, vocabulary size 50265, 0.1 dropout, early stopping, 16 attention heads, 12 layers with feed forward layer dimension set as 4096 in both decoder and encoder.

\noindent Quantized Transformer and Copycat models are trained for 5 epochs.

\section*{Appendix C}
\subsection*{Cluster examples and summaries from the MOS Corpus}

\begin{multicols*}{2}
\label{Appen_C}
\begin{table*}[]
    \centering
    \tiny
    \begin{tabular}{l}
         \toprule
         \multicolumn{1}{c}{\textbf{Tweet cluster fragment for keyword "CDC"}}  \\
         \midrule
\multirow{1}{\textwidth}{Gosh i hope these cases are used for the negligent homicide class action suit that’s being constructed against trump. cdc warns against drinking hand sanitizer amid reports of deaths}\\
\hline
the cdc has also declared, \"being stupid is hazardous to your health.\" URLLINK\\
\hline
cdc warning!  do not drink hand sanitizer! what the hell! people be idiots!\\
\hline
cdc warns against drinking hand sanitizer amid reports of deaths seriously omg?! \\
\hline
if the cdc has to put out a health bulletin to inform people not to try drinking hand sanitizers, how stupid are those people?\\
\hline
\multirow{2}{\textwidth}{from the "if you had any doubt" department: the cdc is alerting your fellow americans not to drink hand sanitizer. obviously more than a couple of people have had to be treated for it.  I wonder were they poisoned in the womb, too many concussions, mt. dew in their milk bottle when they were babies?}\\
\\
\hline
oh my...the cdc actually had to warn people not to drink hand sanitizer. only under a trump presidency have people acted so stupidly.\\
\hline
\multirow{2}{\textwidth}{@realdonaldtrump you should try drinking the hand sanitizer. After your ridiculous suggestion to inject disinfectants, people have decided to drink it and are dying. CDC now issued a warning not to drink it. since u don’t believe anything the scientists say go ahead and drink it. First get kids out of cages}
\\
\\
\hline
@USER i think this actually speaks more to the stupidity of the cdc.\\
\hline
@USER trump is in control of the cdc. don't believe a single word that they are saying\\
\hline

this is sadly what happens when you put an idiot like @realdonaldtrump in the white house...people had seizures, lost vision and dead after drinking hand sanitizer, cdc warns URLLINK\\
\hline
@cdcgov @usfda @USER is it really necessary to tell people not to ingest hand sanitizer????? jesus\\
\hline
cdc gives out warning that people are going blind and are dying from drinking hand sanitizer.  $\#$darwinism URLLINK\\

      \midrule
      \textbf{Human Written Summary}\\
      \multirow{3}{\textwidth}{\textbf{Main Story}: CDC warns people not to drink hand sanitiser after people fall ill and die after doing so. \textbf{Majority Opinion}: The majority are shocked by the stupidity of people thinking it's OK to drink hand sanitiser. \textbf{Minority Opinions}: A minority blame the Trump administration for putting forward the idea of people disinfecting their insides. A minority think the warning is an act of stupidity by the CDC.}
      \\
      \\
      \\

      \midrule
     
      \multirow{2}{\textwidth}{ \textbf{Lexrank Summary}: cdc just issued a warning not to drink it. the cdc just issued an official warning to never drink hand sanitizer and i just.... people are having seizures, losing vision and dying after drinking hand sanitizer, cdc warns  via @yahoo. if you need a warning from the cdc not to drink hand sanitizer... please go ahead and drink hand sanitizer.}\\
      \\
    
      \midrule
      \multirow{2}{\textwidth}{ \textbf{SummPip Summary}: people have decided to drink it and are dying after swallowing hand sanitizer ... children masks .   people are having seizures , losing vision and dying after drinking hand sanitizer , cdc warns via @yahoo .}\\
      \\
      \midrule
    
      \multirow{2}{\textwidth}{  \textbf{BART Summary}: People are getting sick and even dying after swallowing hand sanitizer, the us centers for disease control and prevention said wednesday. four died and others have suffered impaired vision or seizures, the cdc says. The cdc has also declared, "being stupid is hazardous to your health" @realdonaldtrump you}\\
    
      \\
      \midrule
      \multirow{2}{\textwidth}{\textbf{BART FT $10\%$ Summary}: People are dying and even dying after drinking hand sanitizer, the Centers for Disease Control warns. The majority are shocked and appalled by the CDC's warning against people drinking hand Sanitizer. A minority believe the CDC is exaggerating the danger of drinking hand Sanitizinger to distract people from the}\\
      \\
     
      \midrule
      \multirow{2}{\textwidth}{\textbf{BART FT $70\%$ Summary}: CDC warns that people are dying and losing vision and blindness after drinking hand sanitiser. The majority believe that people are extremely stupid for even thinking about doing this, and it should not be allowed to happen. A minority do not believe the CDC's warning and think it is not necessary to take any action}\\
      \\
     
      \bottomrule
    \end{tabular}
    \caption{Example of excerpt from tweet cluster, human summary and best generated summary candidates.}
    \label{tab:appen_example1}
\end{table*}

\begin{table*}[h!]
    \centering
    \tiny
    \begin{tabular}{l}
         \toprule
         \multicolumn{1}{c}{\textbf{Tweet cluster for keyword "mental health"}}  \\
         \midrule
    A 'landmark moment'? Nick Clegg (Lib Dems) promise to put mental health on par with physical \#health URLLINK $\#$inclusion $\#$care\\
    \hline
      All of a sudden, Nick Clegg is concerned about people with mental health issues. Nothing at all to do with trying to win voters and save his job.\\
      \hline
      Delighted that nick is finally doing something about mental health in our nhs\\
      \hline
      Nick Clegg promises 'dignity and respect' in NHS mental health treatment video URLLINK | Guardian\\
      \hline
      I have been hearing very positive noises on the radio today from Lib Dems re: mental health treatment. Certainly long overdue but great to hear!\\
      \hline
      But if you are patting Nick Clegg on the back for new mental health reforms, consider this:\\
      \hline
      Mate, Clegg could have stood up to Cameron before his harmful reductive mental health policies got implemented.\\
      \hline
      Awesome that Clegg highlighted mental health to rapturous applause, but sure he did that with tuition fees once.\\
      \hline
      .\@nickclegg speech \#libdemconf focusing on mental health was cool. Araith Nick Clegg yn canolpwyntio ar iechyd meddyliol yn wych.\\
      \hline
      Nick Clegg's pandering towards the treatment of mental health illness is kinda sad and pathetic$\#$hecantbuyavote\\
      \hline
      One immediate victory of Clegg's speech; putting mental health issues on the agenda and in the media. $\#$ldconf $\#$bbcnews\\
      \hline
LibDems are back to promising the unachievable because they know they’re safe away from power. Shame because mental health is in dire state.\\
\hline
His position in government could have been used to stop the reductive mental health reforms Cameron put in years back. Did he? no.\\

      \midrule
    \textbf{Human Written Summary}\\
      \multirow{2}{\textwidth}{\textbf{Main Story}: Nick Clegg promises to focus on mental health provision in the NHS. \textbf{Minority Opinions}: Some Twitter users are pleased something is 'finally' being done about it and that it is great, it is highlighting mental health. Others are asking why he didn't do it when he was in power and say that Clegg is doing it for personal gain.}\\
      \\
    
      \midrule
     
      \multirow{3}{\textwidth}{ \textbf{Lexrank Summary}: Nick Clegg promises 'dignity and respect' in NHS mental health treatment video  Speaking before his speech to the... Been hearing very positive noises on the radio today from Lib Dems re: mental health treatment. One immediate success of Clegg's speech; getting mental health issues on the agenda and in the media. \@nickclegg a six week wait for mental health related treatment, but didn't hear how you'll resource the \#NHS to achieve the needed care!}\\
      \\
      \\
    \midrule
    
      \multirow{3}{\textwidth}{  \textbf{SummPip Summary}: happy about nick clegg could have been used to stop the reductive mental health treatment . but if you are patting nick clegg is all of a sudden concerned about people with mental health issues . nick clegg promises ' dignity and respect ' in nhs mental health treatment video speaking before his speech to the ... been hearing very positive noises on the radio today from lib dems re : mental health treatment .}\\
      \\
      \\
    \midrule
    
      \multirow{2}{\textwidth}{  \textbf{BART Summary}: Lib Dems promise to put mental health on par with physical health. Nick Clegg promises 'dignity and respect' in NHS mental health treatment video. But if you are patting Nick Clegg on the back for new mental health reforms, consider this: Feeling blessed, trying to eradicate mental health stigma and getting lifetime opportunities}\\
      \\
     
      \midrule
      \multirow{2}{\textwidth}{\textbf{BART FT $10\%$ Summary}:Lib Dem Nick Clegg makes a speech about mental health in the NHS. The majority are pleased that the Lib Dem leader is trying to tackle the stigma attached to mental health. A minority are disappointed that he is pandering to the far right and anti-gay groups. A minority believe he is setting us up for a}\\
      \\
     
      \midrule
      \multirow{2}{\textwidth}{\textbf{BART FT $70\%$ Summary}: Lib Dem leader Nick Clegg makes a speech about putting mental health on a par with physical health in the manifesto. The majority are pleased that Nick Clegg is taking a lead on mental health and saying that mental health needs to be treated with dignity and respect. A minority are dismayed by Nick Clegg }\\
      \\
     
      \bottomrule
    \end{tabular}
    \caption{Example of excerpt from tweet cluster 2, human summary and best generated summary candidates.}
    \label{tab:appen_example2}
\end{table*}
\label{Appen_C}
\end{multicols*}
\end{document}